\pdfoutput=1
\documentclass[12pt]{iopart}
\usepackage{array} 

\usepackage[colorlinks=true,
			citecolor=blue,
			linkcolor=blue,
			urlcolor=blue]{hyperref} 
\usepackage[capitalise]{cleveref}
\usepackage{tabulary,booktabs}
\usepackage[pdftex]{graphicx}

\usepackage[backend=biber,
			giveninits=true,
			maxbibnames=3,
			style=ieee,
			citestyle=numeric-comp, 
			url=false,
			doi=false,
			eprint=false,
			hyperref=true,
			isbn=false,
			date=year
		]{biblatex}
\usepackage{color}
\usepackage{hyphenat}
\begin{document}

\title{Layered Chain-of-Thought Prompting for Multi-Agent LLM Systems:
\\
\large A Comprehensive Approach to Explainable Large Language Models}

\author{Manish Sanwal}

\address{Engineering Department, News Corporation, New York, USA}
\ead{msanwal@newscorp.com}

\begin{abstract}
Large Language Models (LLMs) leverage chain-of-thought (CoT) prompting to provide step-by-step rationales, improving performance on complex tasks. 
Despite its benefits, vanilla CoT often fails to fully verify intermediate inferences and can produce misleading explanations. 
In this work, we propose Layered Chain-of-Thought (Layered-CoT) Prompting, a novel framework that systematically segments the reasoning process into multiple layers, each subjected to external checks and optional user feedback. 
We expand on the key concepts, present three scenarios---medical triage, financial risk assessment, and agile engineering---and demonstrate how Layered-CoT surpasses vanilla CoT in terms of transparency, correctness, and user engagement.
By integrating references from recent arXiv papers on interactive explainability, multi-agent frameworks, and agent-based collaboration, we illustrate how Layered-CoT paves the way for more reliable and grounded explanations in high-stakes domains.
\end{abstract}

%
%
%
%
%

\section{Introduction}

\subsection{Motivation} \label{sec:motivation}

Large Language Models have achieved remarkable success across domains such as question answering, machine translation, and code generation, with widely known architectures like GPT, T5 \cite{wu2024thinking, zhao2024explainability}. A central innovation in recent years is chain-of-thought prompting, wherein an LLM produces a series of intermediate reasoning steps in natural language, often improving both accuracy and interpretability \cite{cambria2024xai,mumuni2025explainable}. However, vanilla chain-of-thought (CoT) methods are prone to two main shortcomings:
\begin{enumerate}
    \item  Unverified Reasoning: The intermediate steps may be plausible but contain hidden contradictions or factual inaccuracies, particularly in specialized fields (e.g., medical, legal, or technical).
    \item Limited User Interaction: The typical CoT pipeline lacks a mechanism for iterative revision or external cross-checking, making it difficult to correct subtle errors once the chain-of-thought is generated \cite{krause2024data,mavrepis2024xai}.
\end{enumerate}
These limitations can have serious consequences in high-stakes applications---for example, in healthcare, a slight misdiagnosis may produce unsafe medical recommendations \cite{hsieh2024comprehensive}. Similarly, in agile software development contexts, overlooked dependencies or flawed logic in LLM-generated code can slow development cycles and introduce latent bugs \cite{manish2024autonomous}. The quest for more reliable, multi-layered explanations that validate partial conclusions has thus become urgent \cite{sanwal2024evaluating,ehsan2024explainable}.

\subsection{Our Contribution: Layered CoT} \label{sec:ourcont}

We propose Layered Chain-of-Thought Prompting, a method that subdivides a model’s reasoning into discrete layers (or blocks). Each layer’s partial output is cross-verified, either via external knowledge sources (domain databases, knowledge graphs) or by user feedback (domain experts, iterative dialogue). By enforcing verification at each layer rather than solely at the end, we improve the faithfulness, consistency, and interactivity of explanations:
\begin{itemize}
    \item Faithfulness: Each partial step can be confirmed or refuted before the model builds on a faulty conclusion.
    \item Consistency: Contradictions are caught early, preventing error propagation.
    \item Interactivity: Users can inject clarifications, constraints, or additional data during the multi-layer reasoning process, enhancing overall trust and usability.
\end{itemize}

Moreover, we show how multi-agent LLM architectures---where specialized agent-models coordinate tasks such as cross-checking, knowledge retrieval, and user interaction---can further strengthen the verification and correction steps in Layered-CoT. We demonstrate through expanded examples how Layered-CoT outperforms vanilla CoT in complex scenarios. We also situate our framework within the broader context of interactive XAI (Explainable AI) approaches \cite{cambria2024xai,krause2024data,mavrepis2024xai,hsieh2024comprehensive}, multi-agent LLM designs \cite{mumuni2025explainable,hsieh2024comprehensive}, and chain-of-thought verification strategies \cite{sanwal2024evaluating,ehsan2024explainable}.

\section{Related Work}

\subsection{Vanilla Chain-of-Thought Prompting}

Chain-of-thought prompting encourages LLMs to produce a textual rationale before stating the final answer \cite{wu2024thinking,zhao2024explainability}. While CoT often significantly enhances model performance on math word problems, reasoning puzzles, or structured QA tasks, it suffers from:
\begin{itemize}
    \item Over-Reliance on Model Confidence: Models might generate plausible-sounding but factually incorrect rationales, misleading end-users \cite{cambria2024xai,mavrepis2024xai}.
    \item Static Explanation Flow: Once the chain-of-thought is produced, there is limited scope for mid-course corrections \cite{krause2024data}.
\end{itemize}

\subsection{Interactive and Contrastive Explanations}

Various studies incorporate contrastive sets and user-in-the-loop techniques to refine or verify model outputs \cite{cambria2024xai,hsieh2024comprehensive,sanwal2024evaluating}. For instance, contrast sets help highlight minimal input perturbations that significantly alter outputs, exposing potential vulnerabilities. User-in-the-loop systems allow domain experts to annotate or correct partial model outputs, thereby iteratively guiding the reasoning process \cite{krause2024data,mavrepis2024xai}. These approaches underscore the importance of iterative refinement but rarely structure the entire chain-of-thought into layers that systematically incorporate external checks.

\subsection{Multi-Agent Approaches and Knowledge Graph Verification}

Research on multi-agent LLM frameworks reveals how separate agent-models can collaborate or cross-check one another’s reasoning \cite{mumuni2025explainable,manish2024autonomous}. For example, one agent might specialize in retrieving authoritative domain information, another in summarizing partial chain-of-thought steps, and yet another in fact-checking or contradiction detection. Additionally, knowledge-graph-based verification can detect factual inconsistencies in the generated chain-of-thought \cite{mumuni2025explainable}. Layered-CoT aligns well with these multi-agent concepts: each layer can be overseen by a specialized agent or a knowledge retrieval module, ensuring partial solutions remain grounded in authoritative data.

\subsection{Agent-Based Collaboration for Layered-CoT}

While Layered-CoT can be implemented with a single LLM plus external checks, a multi-agent pipeline can further subdivide responsibilities:
\begin{itemize}
    \item A Reasoning Agent generates the partial chain-of-thought for each layer.
    \item A Verification Agent cross-checks each partial solution, either via knowledge graphs or external resources.
    \item A User-Interaction Agent collects additional clarifications or constraints from domain experts.
\end{itemize}

This agent-based decomposition helps localize errors and ensures that contradictory or incomplete information can be challenged before polluting subsequent layers. Beyond improving factual accuracy, such a specialization of roles can also optimize computational resources, as each agent can be a smaller model or a rules-based system designed for its specific task.

\subsection{Positioning Our Work}
Layered-CoT extends the above efforts by segmenting reasoning into discrete, verifiable layers. This complements prior attempts at interactive chain-of-thought or knowledge-driven LLMs, offering a methodical pipeline for error interception and collaborative verification. When integrated with multi-agent systems, our approach allows each layer to be validated by dedicated agents, fostering a robust synergy between structured reasoning and distributed agent specialization.
 
\section{The Layered Chain-of-Thought Prompting Framework}

\subsection{Overview}

Layered-CoT structures the reasoning process into layers (Figure 1), each tackling a coherent sub-problem or partial objective. Between layers, the model checks the partial conclusion using external resources (e.g., domain-specific databases, search engines, private knowledge graphs) or human feedback (domain experts). In a multi-agent system, each of these checks can be performed by a specialized agent, further improving transparency and resilience.

\begin{enumerate}
    \item Sub-Problem Identification
    \begin{itemize}
        \item The model (or user) segments the overarching question into sub-questions or tasks (Layer 1, Layer 2, etc.).
    \end{itemize}
    \item Partial Reasoning
    \begin{itemize}
    \item The model proposes a partial chain-of-thought with limited scope.
    \end{itemize}
    \item Verification
    \begin{itemize}
        \item The partial solution is tested or ``cross-verified" via external knowledge or user scrutiny.
        \item In multi-agent setups, a Verification Agent or specialized domain agent can perform this step.
    \end{itemize}
    \item Refinement
    \begin{itemize}
        \item If inconsistencies or gaps are found, the model updates the partial chain-of-thought accordingly.
    \end{itemize}	
    \item Progression
    \begin{itemize}
        \item Once validated, the model proceeds to the next layer until reaching the final layer.
    \end{itemize}
\end{enumerate}

\subsection{Example Workflow}
Prompt: \texttt{"Is Algorithm X suitable for real-time agile software development in a distributed environment?"}
\begin{enumerate}
    \item Layer 1
    \begin{itemize}
        \item Sub-Problem: Evaluate Algorithm X’s time complexity and data synchronization overhead.
        \item Partial CoT: \texttt{"Algorithm X operates at roughly $O(n \log n)$. Sync overhead might be minimal, but needs confirmation."}
        \item Verification Step: Query a performance database, or ask a domain expert. (In a multi-agent scenario, a Knowledge-Retrieval Agent performs the database query; a Verification Agent integrates the new info.)
        \item Refinement: If the overhead is substantial in distributed settings, correct the partial conclusion.
    \end{itemize}
    \item Layer 2
    \begin{itemize}
        \item Sub-Problem: Check concurrency safety.
        \item Partial CoT: \texttt{"Algorithm X ensures concurrency with built-in locking. Might require additional libraries."}
        \item Verification Step: Cross-reference official library documentation or user inputs.
        \item Refinement: Possibly revise to mention recommended concurrency patterns.
    \end{itemize}
    \item Layer 3
    \begin{itemize}
        \item Integration: Combine validated insights from the first two layers into a final answer.
        \item \texttt{"Yes, Algorithm X can be adapted for real-time agile dev. However, concurrency libraries are essential for safe distributed operations."}
    \end{itemize}
\end{enumerate}

By layering partial solutions, the approach intercepts mistakes that might otherwise remain hidden in a single, unverified chain-of-thought. In a multi-agent version of this workflow, specialized agents handle each layer’s retrieval, verification, and user-interaction tasks---further ensuring that each step is robustly checked.
 
\section{Expanded Examples}

We illustrate vanilla CoT versus Layered-CoT across three typical domains: medical triage, financial risk assessment, and agile engineering. Each example shows how Layered-CoT integrates external data checks and user feedback at each intermediate reasoning step to refine the final conclusion. We also note where agent specialization can play a role.

\subsection{Medical Triage Scenario}
User Query: \texttt{"I have had a fever, sore throat, and fatigue for 8 days. Do I need to see a doctor?"}
\begin{itemize}
    \item Vanilla CoT
    \begin{itemize}
        \item CoT Output: \texttt{"These symptoms often suggest a common viral infection. Generally, rest and hydration suffice. If symptoms persist beyond 10 days, see a doctor."}
        \item Potential Issue: The model overlooks factors such as patient age, comorbidities, or local disease outbreaks that might necessitate earlier intervention.
    \end{itemize}
    \item Layered-CoT
    \begin{enumerate}
        \item Layer 1 --- Patient Profile \& Basic Symptoms
        \begin{itemize}
            \item Partial Reasoning: \texttt{"Patient has fever, sore throat, fatigue for 8 days. Check age, region’s infection stats."}
            \item Verification: System consults a local health database to identify high rates of strep throat. (A Health-Data Agent might automate this check.)
        \end{itemize}
        \item Layer 2 --- Risk Factors \& Recommendations
        \begin{itemize}
            \item Partial Reasoning: \texttt{"Possible strep infection. Seniors or immunocompromised patients should seek medical advice immediately."}
            \item Verification: If user data suggests vulnerability, the system updates the recommendation.
        \end{itemize}
    \end{enumerate}
    \begin{itemize}
    \item Outcome: \texttt{"Because local strep cases are high, scheduling a doctor’s visit is advised, especially for at-risk individuals."}
\end{itemize}
\end{itemize}

By checking local data at each step---and potentially leveraging specialized medical agents---Layered-CoT offers more robust guidance, minimizing the risk of overlooked factors.

\subsection{Financial Risk Assessment}
User Query: \texttt{"Is investing \$10,000 in Company X a good idea, given they claim rapid growth in solar panels?"}
\begin{itemize}
    \item Vanilla CoT
    \begin{itemize}
        \item CoT Output: \texttt{"Solar panel demand is rising globally, and Company X is well-positioned. Investment could yield good returns. Proceed if you have spare capital."}
        \item Potential Issue: The model fails to examine the company’s actual financials or potential patent disputes. The conclusion remains largely speculative.
    \end{itemize}
    \item Layered-CoT
    \begin{enumerate}
        \item Layer 1 --- Company Fundamentals
        \begin{itemize}
            \item Partial Reasoning: \texttt{"Company X is in solar manufacturing. Let’s check historical revenue and known patent status."}
            \item Verification: Access a financial database, discovering that the company’s patent is still pending. (A Patent-Agent or Finance-Agent can handle specialized data queries.)
        \end{itemize}
        \item Layer 2 --- Market and Competitive Landscape
        \begin{itemize}
            \item Partial Reasoning: \texttt{"Industry competition is high, and Company X faces rivalry from established brands. Factor in user’s risk tolerance."}
            \item Verification: Cross-check sector reports; if the user is risk-averse, highlight diversification.
        \end{itemize}
    \end{enumerate}
    \begin{itemize}
    \item Outcome: \texttt{"While demand is real, patent uncertainty poses significant risk. Moderate or cautious investment is suggested, pending patent approval."}
\end{itemize}
\end{itemize}

Layered-CoT ensures partial claims---like patent status and competitor analysis---are validated, providing more nuanced advice. In a multi-agent pipeline, specialized Finance Agents handle the data retrieval, while a Risk-Evaluation Agent aligns the advice with the user’s risk tolerance.

\subsection{Agile Engineering Scenario}
User Query: \texttt{"Our software team is expanding rapidly with members across different time zones. Which agile methodology should we adopt to handle frequent requirement changes and ensure timely releases?"}

\begin{itemize}
    \item Vanilla CoT
    \begin{itemize}
        \item CoT Output: \texttt{"Scrum is a popular agile method featuring sprints, daily stand-ups, and iterative planning. It generally improves collaboration and handles changing requirements well."}
        \item Potential Issue: The model may overlook complexities such as distributed time zones, diverse skill sets, and integration challenges with existing DevOps pipelines.
    \end{itemize}
    \item Layered-CoT
    \begin{enumerate}
        \item Layer 1 --- Team Structure \& Time Zones
        \begin{itemize}
            \item Partial Reasoning: \texttt{"Developers span multiple continents, potentially complicating daily stand-ups. Evaluate asynchronous communication tools."}
            \item Verification: Consult known best practices for distributed agile teams, confirm feasibility of daily synchronous meetings.
        \end{itemize}
        \item Layer 2 --- Requirement Volatility \& Tech Stack
        \begin{itemize}
            \item Partial Reasoning: \texttt{"Scrum handles frequent requirement changes, but large user stories might benefit from Kanban’s continuous flow. Check integration with existing DevOps pipelines."}
            \item Verification: Cross-reference pipeline documentation; refine suggestions for backlog management.
        \end{itemize}
    \end{enumerate}
    \begin{itemize}
    \item Outcome: \texttt{"A hybrid approach combining Scrum’s sprint reviews with Kanban’s flexible backlog is recommended, given distributed teams and frequent requirement changes. Verified against known patterns for asynchronous collaboration."}
\end{itemize}
\end{itemize}

By layering the discussion of team constraints and technical requirements, the final recommendation is more context-aware and practical. In a multi-agent version, a Team-Agent might gather data on time-zone overlaps, while a DevOps-Agent checks pipeline integration feasibility.
 
\section{Comparison and Analysis}

\subsection{Qualitative Contrast}

\begin{table}[t]
\caption{Qualitative Comparison of Vanilla CoT vs. Layered CoT.}
\begin{tabulary}{\linewidth}{|C|C|C|}
\hline
Criterion                  & Vanilla CoT                 & Layered CoT                                     \\ \hline
Interpretability           & Basic textual rationale     & Structured, step-by-step verified   reasoning   \\ \hline
Factual Consistency        & Variable, can be misleading & Higher, due to external checks per layer        \\ \hline
User Engagement            & Limited or post-hoc         & Interactive verification at each major   step   \\ \hline
Use in High-Stakes Domains & Risky, may hide errors      & Safer, early detection of contradictory   logic \\ \hline
\end{tabulary}
\label{tab:tab1}
\end{table}

\Cref{tab:tab1} compares vanilla CoT with Layered-CoT along interpretability, correctness, and user engagement dimensions.

\section{Performance Chart}

\begin{figure}[t]
    \centering
    \includegraphics[width=0.8\linewidth]{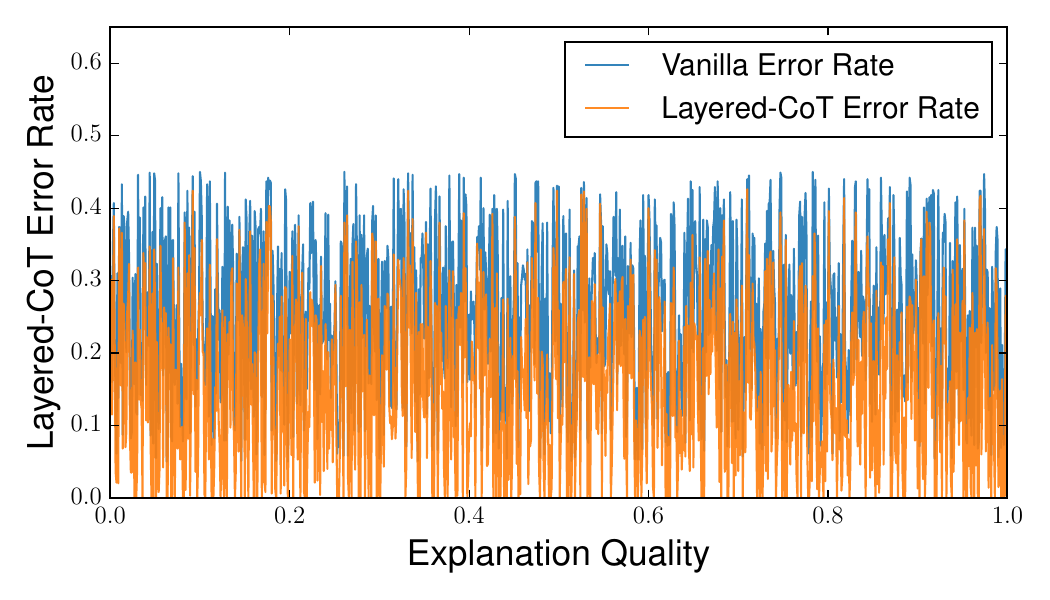}
    \caption{The error rate for the vanilla CoT (blue) and Layered-CoTa (orange) cross explanation qualities.}
    \label{fig:fig1}
\end{figure}

\Cref{fig:fig1} presents a performance chart tested over 1000 tasks—plotting explanation quality against error rate under vanilla CoT vs. Layered-CoT. Layered-CoT typically shows reduced error rates (Y-axis) for a given level of explanation quality (X-axis), especially in multi-step tasks.

\subsection{Implementation Feasibility}
\begin{itemize}
    \item Computational Overhead: Layered-CoT typically requires multiple interactions (one per layer), which can be more expensive in real-time systems. 
    \item External Knowledge Dependencies: Domain data or specialized APIs may be needed to verify partial reasoning.
    \item User Workflow: Where domain expertise is scarce, user-in-the-loop steps must be designed for minimal friction \cite{krause2024data,mavrepis2024xai}.
    \item Agent Coordination: In multi-agent scenarios, an additional coordination mechanism is needed. While this may add complexity, it allows specialized agents to focus on targeted tasks (knowledge retrieval, verification, user interaction).
\end{itemize}

\section{Discussion and Future Directions}
\begin{enumerate}
    \item Scalability and Automation: Implementing partial or automated checks (e.g., web search, knowledge graph lookups) can mitigate the cost of repeated prompts. Agent-based division of labor can reduce the load on a single LLM by delegating specific tasks to simpler models or rule-based systems.
    \item User Studies and Real-World Validation: Human-centered research is needed to measure how end-users perceive layered vs. vanilla explanations—e.g., trust, mental workload, adoption rate \cite{krause2024data,hsieh2024comprehensive}. Multi-agent prototypes should be tested to see whether dividing tasks among agents improves usability.
    \item Multi-Agent Extensions: Layered-CoT pairs well with multi-agent architectures, where each layer can be handled by a specialized agent for domain-specific verification \cite{mumuni2025explainable,manish2024autonomous}. Future work could explore agent orchestration protocols (e.g., dispatching layers to relevant domain agents automatically).
    \item Adaptive Depth: Not all questions require the same number of layers. Future developments could use heuristics or user feedback to dynamically adjust the layering depth. Agents could also negotiate the number of layers based on complexity or risk level.
    \item Cross-Domain Generalizability: We believe the framework can extend to numerous areas, including code generation, geoscience data analysis, or humanitarian response planning. Agent-based specialization—e.g., a geoscience agent or an aid logistics agent—further enhances domain-specific verification.
    \item Agent Coordination and Accountability: As multi-agent LLM systems become more prevalent, ensuring that each agent’s responsibilities are clear and that accountability is maintained across agents will be critical. Layered-CoT offers a natural structure to track how each agent’s outputs factor into the final decision or explanation.
\end{enumerate}
 
\section{Conclusion}

We presented Layered Chain-of-Thought (Layered-CoT) Prompting, an approach that systematically divides the reasoning path of an LLM into verifiable layers. By validating partial conclusions with external sources or user feedback, we address longstanding concerns regarding unverified or misleading rationales in vanilla CoT. Through detailed examples spanning healthcare, finance, and agile engineering, we illustrate the practical benefits of layering, including higher transparency, improved correctness, and heightened user trust.

When combined with multi-agent systems, Layered-CoT becomes even more powerful: specialized agents can handle verification steps, knowledge retrieval, or user interaction, providing a highly modular and resilient pipeline. While Layered-CoT may introduce computational overhead, its promise in high-stakes environments justifies further research into partial automation, multi-agent synergy, and user-focused evaluation. Our hope is that the approach fosters more responsible and robust usage of LLMs, ultimately bridging the gap between high-level performance gains and explainable, verifiable AI reasoning.

\nocite{*} 
\printbibliography

\end{document}